\documentclass[10pt,journal,compsoc]{IEEEtran}
\usepackage{subcaption}
\usepackage{flushend}
\usepackage{graphicx}
\usepackage{hyperref}

\ifCLASSOPTIONcompsoc
  \usepackage[nocompress]{cite}
\else
  \usepackage{cite}
\fi
\ifCLASSINFOpdf
\else
\fi
\begin{document}
%
\title{Multi-view Gradient Consistency for SVBRDF Estimation of Complex Scenes under Natural Illumination}
%
%
%
%

\author{Alen~Joy,~\IEEEmembership{Member,~IEEE,}
        and~Charalambos~Poullis,~\IEEEmembership{Senior~Member,~IEEE}
\IEEEcompsocitemizethanks{\IEEEcompsocthanksitem A. Joy and C. Poullis are with the Immersive and Creative Technologies Lab at the Department
of Computer Science and Software Engineering at Concordia University, Canada.\protect\\
E-mail: alenj445@gmail.com, charalambos@poullis.org}
\thanks{Manuscript received April 19, 2005; revised August 26, 2015.}}

%
%

\markboth{Journal of \LaTeX\ Class Files,~Vol.~14, No.~8, August~2015}%
{Shell \MakeLowercase{\textit{et al.}}: Bare Demo of IEEEtran.cls for Computer Society Journals}
%



\IEEEtitleabstractindextext{%
\begin{abstract}
This paper presents a process for estimating the spatially varying surface reflectance of complex scenes observed under natural illumination. In contrast to previous methods, our process is not limited to scenes viewed under controlled lighting conditions but can handle complex indoor and outdoor scenes viewed under arbitrary illumination conditions. An end-to-end process uses a model of the scene's geometry and several images capturing the scene's surfaces from arbitrary viewpoints and under various natural illumination conditions. We develop a differentiable path tracer that leverages least-square conformal mapping for handling multiple disjoint objects appearing in the scene. We follow a two-step optimization process and introduce a multi-view gradient consistency loss which results in up to 30-50\% improvement in the image reconstruction loss and can further achieve better disentanglement of the diffuse and specular BRDFs compared to other state-of-the-art. We demonstrate the process in real-world indoor and outdoor scenes from images in the wild and show that we can produce realistic renders consistent with actual images using the estimated reflectance properties. Experiments show that our technique produces realistic results for arbitrary outdoor scenes with complex geometry. The source code is publicly available at:  \url{https://gitlab.com/alen.joy/multi-view-gradient-consistency-for-svbrdf-estimation-of-complex-scenes-under-natural-illumination}
%
\end{abstract}

}

\maketitle

\IEEEdisplaynontitleabstractindextext

%
\IEEEpeerreviewmaketitle

\begin{figure*}[!ht]
\centering
 \includegraphics[width=\textwidth]{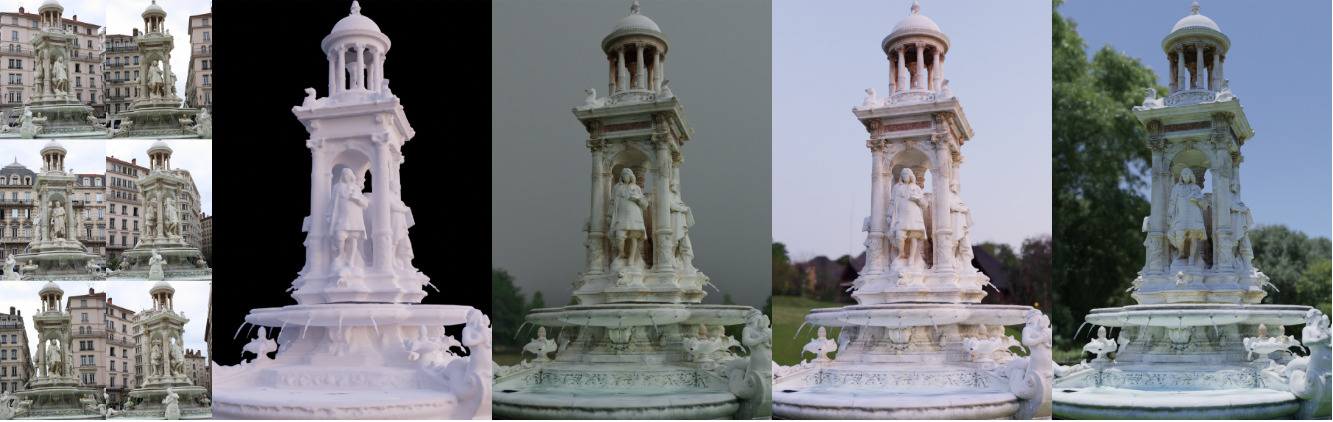}
 \centering
  \caption{We present an end-to-end process for accurately estimating reflectance and re-rendering objects under novel viewpoints and lighting conditions for complex scenes under natural illumination using multi-view images in the wild. We modulate the reconstruction loss by a novel multi-view gradient consistency loss within a two-step optimization process, resulting in faster convergence, improved disentanglement of the reflectance parameters, and RMSE improvements on the reconstruction loss.(first column) multi-view images, (second column) scene's geometry, (columns 3-5) re-rendering under various environment maps.
  }
\label{fig:teaser}
\end{figure*}

\IEEEraisesectionheading{\section{Introduction}\label{sec:introduction}}

%
%
%
%
\IEEEPARstart{T}{his} Rendering photorealistic images is an integral part of computer graphics that involves capturing the scene's geometry and reflectance properties. This becomes inherently difficult in outdoor complex scenes viewed under natural illumination conditions. 

There are many available techniques for capturing the scene's geometry. Some of the most commonly used are LiDAR scanning \cite{bybee2019fusedlidar}, multi-view stereo \cite{schonberger2016structure}, structured-light scanning \cite{gupta2011structured}, shape from shading \cite{yang2018shapefromshading}, and photometric stereo \cite{yvain2016photometricstereo}. Multi-view stereo (MVS) has emerged as the most cost-effective and accurate for outdoor and large-scale scenes, as many recent results attest \cite{Yao_2018_ECCV, Xue_2019_ICCV,Liu_2020_CVPR}.

In addition to the scene's geometry, it is also necessary to have the scene's reflectance properties. That is, how each surface point in the scene reflects light. Similarly, there are many techniques for capturing reflectance properties ranging from directly measuring the reflectance using a gonioreflectometer to estimating the reflectance properties by formulating the problem as inverse rendering. The biggest challenge when using these techniques is scalability to large-scale objects or scenes, and in almost all cases, strong assumptions on the lighting conditions. The most relevant work in terms of the objective and scale is the work of Debevec et al. in \cite{debevec2004estimating}, where they used a tailor-made device to measure the BRDF of four representative areas of an outdoor monument. The measurements took place at night to ensure controlled lighting conditions required for the capture. They recovered spatially varying BRDF based on the assumption that all surface points on the monument's geometry reflect light the same as one of the four measured BRDFs. However, this is a strong assumption, and although it may be justified in the case of a monument made out entirely of the same material as in their case, it does not hold for general outdoor scenes.

More recently,  a series of works on neural re-rendering \cite{yan2018von,chen2019dibrender, Meshry2019Neural, zhang2021image} were proposed as an alternative solution to producing realistic renders under changing lighting and viewpoint. These techniques can produce remarkable results and apply to larger-scale scenes than before; however, at the cost of having minimal control over the output. All information, including the reflectance properties, are embedded in the network with no obvious way to disentangle them, limiting their reusability with other renderers. Moreover, the appearance of the scene is not consistent between renders from novel viewpoints and lighting.

This paper describes a process for estimating spatially varying surface reflectance of complex scenes under natural illumination. We use images in the wild of real-world indoor and outdoor scenes and reconstruct the scene's geometry using multi-view stereo. We model the reflectance properties using a parameterization of the modified Cook-Torrance BRDF. The spatially varying BRDFs are optimized using a differentiable path tracer. Unlike other differentiable renderers that assume a single organic object in the images, leveraging least-square conformal mapping enables us to handle multiple disjoint objects with complex geometry that appear in the scene. To address challenges with disentangling diffuse and specular properties, we follow a two-step optimization process and introduce multi-view gradient consistency loss. 
To summarize, our contributions are:
\begin{itemize}
	\item An end-to-end process for estimating spatially varying BRDF for complex scenes under natural illumination. Unlike existing state-of-the-art our focus is on large-scale outdoor scenes.
	\item A differentiable path tracer that formulates inverse rendering as a two-step optimization and introduces a new multi-view gradient consistency loss. In contrast to existing works that assume a single small-scale synthetic object with simple geometry, we can estimate the reflectance properties of an arbitrary number of disjoint objects with complex geometries appearing in the scene. The two-step process improves the disentanglement of the diffuse and specular reflectance. When combined with the proposed loss term the reconstruction loss is improved by up to 30-50\% for scenes containing non-metallic surfaces, and 5-10\% for scenes containing metallic surfaces.
\end{itemize}

Our technique is tested on a benchmark dataset containing large-scale real-world scenes curated for reconstruction from aerial images with multi-view stereo \cite {yao2020blendedmvs}. We further validate the accuracy of our results by comparing the estimated reflectance properties of synthetic scenes to their ground truth, and a state-of-the-art differentiable path tracer.






\section{Related Work}
Appearance modelling has been a topic of interest to the computer graphics and vision research communities for many years. Many techniques have been proposed for estimating reflectance properties which can be better categorized as procedural-based or deep-learning-based.

Procedural appearance modelling techniques have been around for decades. These techniques most often require specific setups for the acquisition of the appearance of the scene. In \cite{Mcallsiter2002sbrdf}, the per-point reconstruction of reflectance is performed by sampling the complete 4D BRDF function using a gonioreflectometer. Gardner et al. \cite{gardner2003light} proposed a method for estimating the reflectance properties of a surface by passing a linear light source. It attempts to estimate the diffuse and specular colours along with the specular roughness of each surface point by comparing the observed results to a tabulated real reflectance values rendered beforehand. In \cite{Dror2001estimatingsurface}, Dror et al. developed a machine vision system for the estimation of reflectance using statistical features of images. The technique uses SVMs to classify based on the computed statistics. However, the method relies solely on the surface's image; therefore, the results can be ambiguous depending on the illumination and light intensity.

Zickler et al. \cite{Zickler2005color} presented a method of computing a data-dependent rotation of RGB colour space to obtain a photometric reconstruction of the image. This technique does not involve using an explicit reflectance model or a reference object, and the reconstruction is formulated as a scattered interpolation problem. Lombardi and Nishino \cite{Lombardi2012SingleImage} proposed a technique for jointly estimating the material and the light source from a single image by creating a solution space for real-world materials and using a probabilistic approach.
 
While several data-driven approaches have been proposed for obtaining accurate BRDFs, they require a specialized apparatus or laboratory setup which imposes constraints on the shape of the objects and the lighting conditions to be sampled effectively. Furthermore, the lack of tools that can support larger-scale outdoor scenes reduces the efficiency of such techniques.

Recent advancements in deep learning have shown significant performance improvements on both vision and graphics-related tasks.  This, in turn, has also led to the rise of many deep learning-based approaches in appearance modelling, which could tackle some of the challenges posed by the procedural techniques.

Deep learning-based techniques for appearance modelling are most often represented as an end-to-end learned system. In \cite{Li2017Singlephotonet}, the authors presented a solution for estimating SVBRDFs from a single image of a planar surface under unknown illumination. However, due to the lack of control over the lighting, there is no guarantee on obtaining an accurate specular estimation. Using direct lighting can provide more information on the specularity of the surface. Leveraging this, the authors in \cite{li2018materials} proposed a material acquisition technique using images lit by a flashlight to capture high-frequency specular highlights.

Li et al. \cite{li2018learning} demonstrated the recovery of SVBRDF, and complex geometry from a single RGB image illuminated by an environment map and flashlight. This is achieved by training a deep neural network on multiple images under different views and lighting conditions. Meshry et al. \cite{Meshry2019Neural} attempted total scene capture including recording, modelling, and rendering scenes from publicly available photos of landmarks. A neural network was trained to map an initial rendering from scene points to the actual photos.

To enable deep learning models to understand image formation or rendering a 3D scene has led to the development of differentiable renderers. Differentiable renderers constitute a class of techniques that handle the integration of the rendering process into deep learning models for end-to-end optimization by back-propagating gradients from rendered images to scene information.  Chen et al. \cite{chen2019dibrender} developed a differentiable interpolation based renderer, which attempts to recover the 3D shape and texture information. This is done by treating the foreground rasterization process as an interpolation of vertex attributes capable of generating realistic images, whose gradients can be back-propagated for optimization. \cite{ravi2020pytorch3d} introduced a library of operators for 3D deep learning including a modular differentiable rendering engine composed of a rasterizer and shader, designed to compute gradients with respect to inputs including camera, textures and lighting.

\cite{NimierDavidVicini2019Mitsuba2} supports a differentiable rendering algorithm capable of computing derivatives of the entire scene for input parameters such as geometry, camera pose, BRDFs, and texture. Combining it with gradient-based optimization, it is capable of solving inverse problems. Li et al. in \cite{Li:2018:DMC}, introduced a differentiable ray tracer for inverse rendering that stochastically computed gradients without approximations. The differentiable ray tracer can also handle secondary lighting effects such as shadows and global illumination. However, in most of these techniques, the scenes contain a single small-scale organic object with simple geometry and are not optimized to obtain the reflectance properties and re-render in new lighting conditions.

In summary, procedural techniques are capable of recording accurate BRDFs under a controlled environment. This is often done with objects with simpler geometry in controlled lighting setups. On the contrary, the recent deep learning techniques are developed end-to-end without much control over how the network interprets the materials, leading to inaccuracies in the recovered material properties, primarily due to entanglement in the BRDF components.


\section{Reflection Model}
\label{sec:reflection_model}

We proceed with the definition of the reflection model. The bidirectional reflectance distribution function (BRDF) is a function that defines the reflectance at a surface point $P$ and is given by $f_{r}(P, \omega_{i}, \omega{o}) = \frac{L(P, \omega_{o})}{L(P, -\omega_{i})cos(\phi)m(\Omega)}
    \label{eq:brdf}$ where $\omega_i$ and $\omega_o$ are the incoming and the outgoing light directions, $\phi$ is the colatitude of the light source, $\Omega$ represents the solid angle formed by the light source at the surface point $P$, and $m(\Omega)$ specifies its measure. The incoming and outgoing light directions $\omega_i$ and $\omega_o$ are further parameterized by azimuth angle $\phi$ and zenith angle $\theta$ making the BRDF a function of four parameters $f_{r}(\phi_{i}, \theta_{i}, \phi_{o}, \theta_{o})$.

\textbf{Acquisition.} The acquisition of BRDF involves the measurement of the amount of incoming radiance that contributes to the final reflected light at $P$ for all combination of $\omega_{i} \in \textbf{S}^{2}_{-}, \omega_{i} \in \textbf{S}^{2}_{+}$. $\textbf{S}^{2}$ is used to denote the unit sphere in 3D-space, i.e. the set of all possible directions light can flow. $\textbf{S}^{2}_{+}$ includes all the vectors pointing away from the surface, while $\textbf{S}^{2}_{-}$ includes the vectors pointing to the surface.

Acquiring BRDF measurements under a lab setup is a straight-forward process that involves the use of a gonioreflectometer in a dark room. The lack of any additional light sources eliminates interference with the measurements. A spherical arm supports a spotlight that illuminates the sample in the center of the sphere. The position of the spotlight is varied and the amount of light bouncing off the sample will be measured by a sensor.

In the case of large objects located outdoors, capturing BRDF measurements is a difficult and time-consuming task, because of the complex lighting conditions that cannot be controlled. During day, the sun serves as the only light source for the scene, but at night a number of light sources (e.g., moon, light post, flashlights, etc) may contribute to the lighting of the scene. In both cases, controlling the lighting conditions is nearly impossible and obtaining BRDF measurements for all surface points of the object becomes unfeasible unless strong assumptions are made.


In our case, there are additional restrictions that aggravate the difficulties of capturing BRDFs: (i) We do not have access to perform close-up reflectance measurements even though the structures are outdoors, (ii) Images of outdoor scenes are typically in the form of wide-area motion imagery that are captured from an aerial sensor which is orbiting the scene, (iii) Due to the weak-perspective (i.e. the distance of the sensor from the scene is considerably larger than the range of depth within the scene) each scene captures a large physical area, (iv) The image capture is done during the day. For these reasons, we are limited to image-based BRDF acquisition. 

\textbf{Representation.} Assuming that all problems associated with measuring BRDFs have been resolved, the issue of storage and representation remains. Using dense/full BRDF measurements can provide the most accurate results. There is also an option for the BRDF for a particular pair $(\omega_i, \omega_o)$ to be retrieved or interpolated from nearby samples. However, this increases the computational complexity and imposes a significant cost for performing effective sampling during rendering. In this work, we use a combination of phenomenological and physically-based scattering models.

\begin{figure*}[!ht]
\centering
\includegraphics[width=\textwidth]{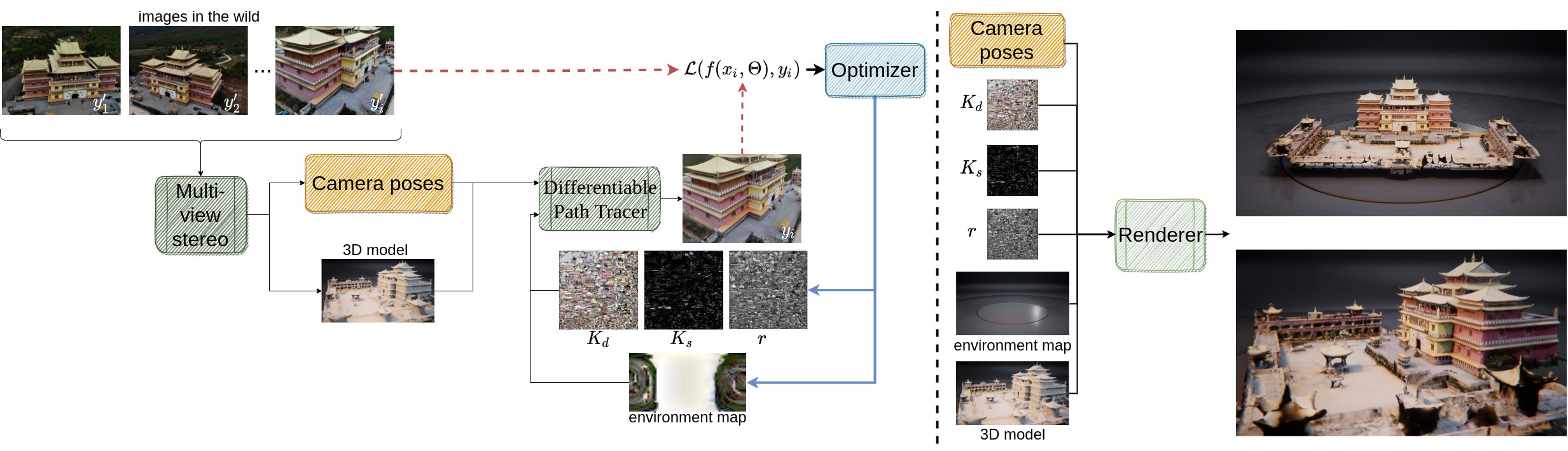}
\caption{We use multi-view images of large-scale scenes of complex geometry under natural illumination to generate a model representing the scene's geometry using multi-view stereo. Given the multi-view images, the model, and camera poses, we present an end-to-end process for accurately estimating reflectance and re-rendering objects under novel viewpoints and lighting conditions. We optimize a reconstruction loss modulated by a novel multi-view gradient consistency loss $\mathcal{L}_{MVCL}$ that enforces a constraint on the optimization parameter space and results in faster convergence, improved disentanglement of the diffuse and specular reflectance, and RMSE improvements on the reconstruction loss of up to 30-50\% for mostly non-metallic scenes and 5-10\% for mostly metallic scenes.}
\label{fig:system_overview}
\end{figure*}

A BRDF approximates the surface's reflective and scattering properties. It should take into account the diffuse reflection formed by scattering the incident light equally in all directions, while also considering the specular reflection. We employ the modified Cook-Torrance BRDF model \cite{cook1982model} to represent the surface properties. This model is a linear blend of a Lambertian reflection model and a micro-facet BRDF. The complete model is given by $f(P,\omega_i,\omega_o) = dR_L(P,\omega_i,\omega_o) + sR_{CT}(P,\omega_i,\omega_o)$
where $d$ and $s$ are the ratio of the incoming light $L_i$ that is reflected $L_o$ for each of the diffuse and specular components, respectively. $R_L$ is the Lambertian reflection model, which is a constant factor denoted as $R_L(P,\omega_i,\omega_o) = \frac{\rho}{\pi}$ where $\rho$ is the albedo. The reflected radiance $L_o$ varies linearly with the incident radiance $L_i$ and is given by $L_o = R_L(P,\omega_i,\omega_o)L_i = \frac{\rho L_i}{\pi}$. The specular term of the BRDF is given by $R_{CT} = \frac{DFG}{4(\omega_o{\cdot}n)(\omega_i{\cdot}n)}$ where $D$ is the Phong distribution function, $F$ is the Fresnel term which is approximated using Schlick's approximation \cite{Schlick94aninexpensive}, and $G$ is a geometric attenuation function approximated with Smith's method with Schlick-GGX. The relation between the reflected radiance $L_o$ and the incoming radiance $L_i$ is given by $L_o = R_{CT}(P,\omega_i,\omega_o)L_i{n}{\cdot}\omega_i{d}\omega_i$.

\section{Method}
\label{sec:method}
We estimate an object's spatially varying BRDF from a set of images capturing the object under unknown lighting conditions. The BRDF is parameterized with three reflectance maps for diffuse, specular and specular roughness. The environment map and the three reflectance maps are estimated with a differentiable path tracer in a two-step inverse rendering optimization, subject to a composite loss function consisting of two terms. The first term measures the per-pixel similarity between the input and rendered images. The second term enforces a multi-view gradient consistency on each pixel, in each map. The formulation of the two-step process combined with the loss function results in the improved disentanglement of the diffuse and specular components, as we present in the subsequent sections. 



\subsection{Differentiable rendering} 
Differentiable rendering is a state-of-the-art technique for solving inverse-rendering problems. The problem is reformulated as finding the optimal parameters $\Theta$, used in the mathematical operations of a function $f$ representing the physical simulation of the path tracing algorithm \cite{renderingeqn}, which given input parameters $x$, minimize a loss function $\mathcal{L}(y, y')$ conditioned on the ground truth image $y$ and the output rendered image $y' = f(x, \Theta)$. The input parameters include the camera pose $x_{cam}$, and the object's geometry $x_{geo}$ i.e. $x = <x_{cam}, x_{geo}>$.

Following the reflection model described in Section \ref{sec:reflection_model}, the optimization parameters $\Theta$ consist of $\theta_{env}$ and the parameterization of the BRDF into the three reflectance maps $< \theta_{diff}, \theta_{spec}, \theta_{rough}>$.
In the context of differentiable rendering, back-propagation calculates the gradients $< \frac{\partial \mathcal{L}}{\partial \theta_{diff}}, \frac{\partial \mathcal{L}}{\partial \theta_{spec}}, \frac{\partial \mathcal{L}}{\partial \theta_{rough}}, \frac{\partial \mathcal{L}}{\partial \theta{env}}>$ which provide a measure of how the loss function $\mathcal{L}$ can be optimized by updating the values of the parameters $\Theta$ towards the gradients' directions.

Differentiable path tracers can simulate photons, produce realistic results, deal with secondary lighting effects such as shadows and indirect light, and recreate outdoor scenes with no visible loss of detail. The results of recent work \cite{zhang2019differential, zhang2020path, loubet2019reparameterizing} demonstrate the effectiveness of differentiable path tracers in generating realistic results. However, these techniques are limited to single, contiguous objects with simple geometries.

Furthermore, the disentanglement of the spatially varying reflectance into diffuse, specular and specular roughness components using differentiable rendering without making assumptions on the object's reflectance or the lighting conditions is an ill-posed problem. The multiple views of the scene's surfaces provide a limited number of measurements of the scene's reflectance that may not capture sufficient outgoing radiance directions to model how the surface reflects light. Using an $L_{2}$-norm on the image similarity between a rendered and actual input image, a differentiable renderer gives the optimal result for the reflectance in terms of the final rendered image. However, the diffuse and specular components are entangled, and although together can provide visually plausible results, they do not match the actual reflectance properties of the scene \cite{luan2021unified}. 

\subsection{Loss function} 
Our process follows a two-step optimization. In the first step, we solve for spatially varying diffuse reflectance and environment map using an $L_{2}$-norm reconstruction loss on the image similarity between a rendered and actual input image given by,
\begin{equation}
	\mathcal{L}_{recon} = \frac{1}{N} \sum_{i=1}^{N} ||y'_{i} - f(x_{i},\Theta) ||_{ _{2}}
	\label{eq.first_step_loss}
\end{equation}
where $f(x_{i},\Theta)$ is the rendered image, $y'_{i}$ is the ground truth image, and $1<i<N$ where N is the number of views. Solving for only diffuse reflectance is based on the observation that the number of non-metallic surfaces in outdoor scenes significantly surpasses that of metallic surfaces. Although this introduces a bias towards non-metallic materials, as we show in Section \ref{sec:experimental_results}, our process can achieve improved disentanglement of the diffuse and specular reflectance even in scenes containing only metallic surfaces. This can be attributed to the more accurate initialization of the diffuse reflectance, which results in the optimization parameters not getting stuck in local minima when minimizing the loss function using all reflectance maps.

The result of the first optimization step provides the initialization for the diffuse reflectance map. During the second optimization step, the three reflectance maps are optimized together with the environment map, subject to the loss function given by,
\begin{equation}
	\mathcal{L} = \frac{1}{N} \sum_{i=1}^{N} (\mathcal{L}_{recon}^{i} \times e^{\mathcal{L}_{MVCL}^{i}})
	\label{eq.second_step_loss}
\end{equation}
where $\mathcal{L}_{recon}$ is the reconstruction loss similar to the first step, and $\mathcal{L}_{MVCL}$ is a multi-view gradient consistency loss which is a function of the variances of the gradients of each surface point $p$ in the scene $S$ visible in $N$ views and is given by,
\begin{equation}
	\mathcal{L}_{MVCL} = \sum_{diff} \sum_{p}^{S} \left[ \frac{1}{N-1} \sum_{i=1}^{N} \left[\frac{\partial \mathcal{L}_{recon}^{i}}{\partial \theta_{diff}} - \frac{\partial \mathcal{L}_{recon}^{\mu}}{\partial \theta_{diff}}\right]_{p}^{2} \right]
	\label{eq.variance_step_loss}
\end{equation}
where $\frac{\partial \mathcal{L}_{recon}^{\mu}}{\partial \theta_{diff}}$ is the sample mean gradient for point $p\in S$.

We argue that the $\mathcal{L}_{MVCL}$ term plays a key role in our loss. Consider a surface point $p\in S$ that is visible in $N$ views. The gradients for point $p$, for the diffuse reflectance, are $(\frac{\partial \mathcal{L}_{recon}}{\theta_{diff}})_{1}, (\frac{\partial \mathcal{L}_{recon}}{\theta_{diff}})_{2}, (\frac{\partial \mathcal{L}_{recon}}{\theta_{diff}})_{i}, ..., (\frac{\partial \mathcal{L}_{recon}}{\theta_{diff}})_{N}$, each calculated during back-propagation for every rendered image $y_{i}$ and ground truth $y_{i}'$, where $1<i<N$. In an inverse rendering problem, although the loss for each view will likely evaluate to a different value, the gradients are calculated with respect to the same reflectance parameters and must therefore be the same for each point $p$. Hence, $\mathcal{L}_{MVCL}$ forces the unbiased variance of the gradients to zero and consequently the direction of the gradients to be the same for each point $p\in S$. 

We normalize the gradients to ensure that the variance is bound in the range [0, 1] for every epoch and avoid trivial solutions and vanishing gradients. Hence, in the case where the variance of the gradients is zero the loss function $\mathcal{L}$ reduces to $\mathcal{L}=\mathcal{L}_{recon}$, otherwise it increases exponentially based on $\mathcal{L}_{MVCL}$ with a maximum value of $\mathcal{L}=\mathcal{L}_{recon}\times e^{1}$. The multi-view gradient consistency loss results in improvements of up to 30-50\% on the final reconstruction loss in scenes with mostly non-metallic surfaces, and 5-10\% for mostly metallic surfaces as presented later in Section \ref{sec:experimental_results}. Furthermore, it provides faster convergence and optimized reflectance with reduced light effects including shadows, attributed to the improved disentanglement of the reflectance parameters.

\section{Experimental Results}
\label{sec:experimental_results}
We first present experiments on the effectiveness of the $\mathcal{L}_{MVCL}$ and its effect on the disentanglement of the diffuse and specular reflectance. Next, we present the results on large-scale outdoor scenes containing multiple disjoint objects with complex geometry, and experiments on determining the optimal resolution for the reflectance maps.

\subsection{Implementation} 
We initialize the values of the diffuse and specular reflectance to $\theta_{diff} = \{0\}, \theta_{spec} = \{0\}$ and the specular roughness to $\theta_{rough} = \{0.5\}$. Although there is no upper bound on the resolution of  $\theta_{diff}, \theta_{spec}, \theta_{rough}$, our experiments with $512^2, 1024^2, 2048^2, 4096^2$ show that a resolution of more than $2048^2$ provides negligible improvement on the loss at a significant computational cost. Therefore, for all reported experiments the resolution is fixed to $2048\times2048$. An environment map is used to estimate lighting conditions captured in the input images and is initialized to $\theta_{env} = \{0\}$. We experimentally determine that a reduced resolution of $32\times64$ for the environment map provides the best trade-off between accuracy and computational efficiency of the optimization. All reported results were computed on a workstation with Intel i9 processor and a 12GB Nvidia RTX 2080 GPU.

\begin{figure}[!ht]
\hspace{-15pt}
\includegraphics[width=0.5\textwidth]{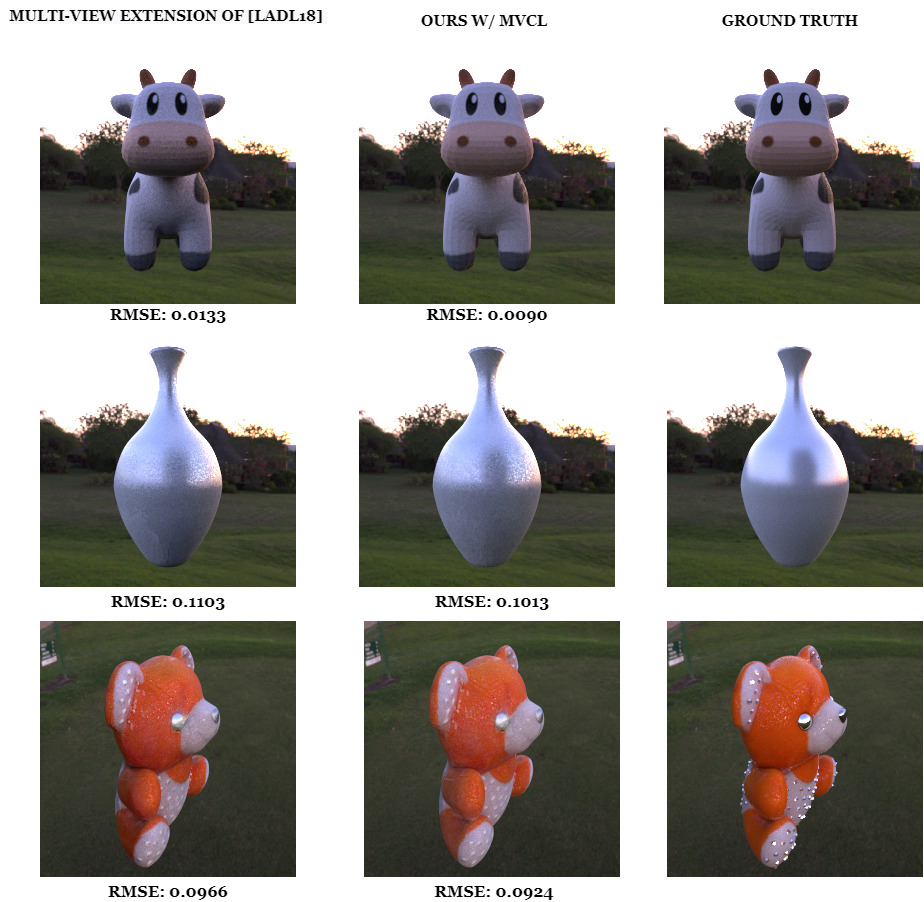}
\caption{Effectiveness of $\mathcal{L}_{MVCL}$. Left column: multi-view extension of \cite{Li:2018:DMC} with conformal mapping. Middle column: our technique with $\mathcal{L}_{MVCL}$. Right column: ground truth. Top row: Perfectly diffuse. Middle row: Highly specular. Bottom row: Diffuse + Specular. The RMSE value for the diffuse cow in top row is lower by 32\%. For the highly specular vase in middle row the RMSE value is 9\% lower and the teddy-bear with combined material properties shows a 5\% lower RMSE value. All synthetic examples shown in the paper use 48 multi-view images. }
\label{fig:effects_of_mvcl}
\end{figure}

\subsection{Effectiveness of $\mathcal{L}_{MVCL}$} 
The proposed multi-view gradient consistency loss $\mathcal{L}_{MVCL}$ plays a key role in the optimization. We demonstrate this by comparing it with the state-of-the-art differentiable path tracer in \cite{Li:2018:DMC} using synthetic objects for which ground truth reflectance is available. To ensure a fair comparison, we extend \cite{Li:2018:DMC} to handle multi-view images, and use least-squares conformal mapping instead of the angle-based flattening to map the reflectance maps to the object's geometry. 

We use three different synthetic datasets that provide ground truth for different material properties. Figure \ref{fig:effects_of_mvcl} shows the results of this experiment. The left column shows the result of the multi-view extension of \cite{Li:2018:DMC}, the middle column shows the result using the multi-view gradient consistency loss, and the third column the ground truth. In all three experiments, the RMSE when using $\mathcal{L}_{MVCL}$ is lower. For the diffuse cow in the top row, the RMSE showed a 32\% decrease, while the highly specular vase in the middle row, and the teddy-bear with combined material in the bottom row, showed 9\% and 5\% lower RMSE values, respectively. The calculation of RMSE includes only pixels corresponding to surface points of the object, and not the environment map. 

\begin{figure}[!ht]
\centering
\includegraphics[width=0.5\textwidth]{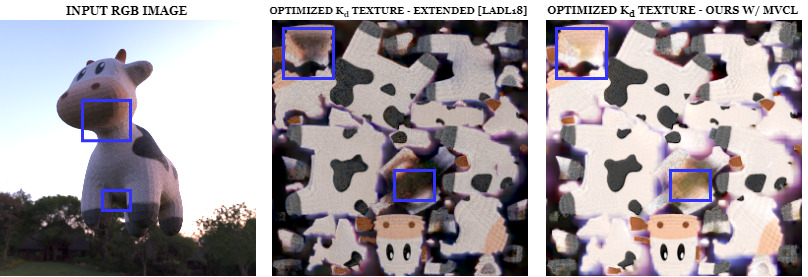}
\caption{The blue rectangles show areas with fine details that were missed (middle column) with \cite{Li:2018:DMC} and captured by our technique(right column). The example shows the diffuse reflectance only.}
\label{fig:capturing_hidden_features}
\end{figure}

Furthermore, using the multi-view gradient consistency loss captures view-dependent details and features obstructed by self-occlusion. Figure \ref{fig:capturing_hidden_features} shows example patches marked in blue rectangles indicating missed details by \cite{Li:2018:DMC} (middle column) that are captured using our technique in the right column. In this example, the lack of shadows is clear evidence that lighting information has been decoupled from the diffuse reflectance.

\subsection{Effects on entanglement} 
A limitation of inverse rendering is the entanglement of the optimized parameters $\theta_{diff}, \theta_{spec}, \theta_{rough}, \theta_{env}$ and in particular the diffuse and specular reflectance. Re-rendering a scene with entangled reflectance parameters can result in a visually identical render to the ground truth image. However, this entanglement creates an inherent dependency between the parameters, which can only produce realistic results if used in unison. Changing any reflectance parameter will cause visual artifacts and a sharp increase in reconstruction loss. 

\begin{figure}[!ht]
\centering
\includegraphics[width=0.5\textwidth]{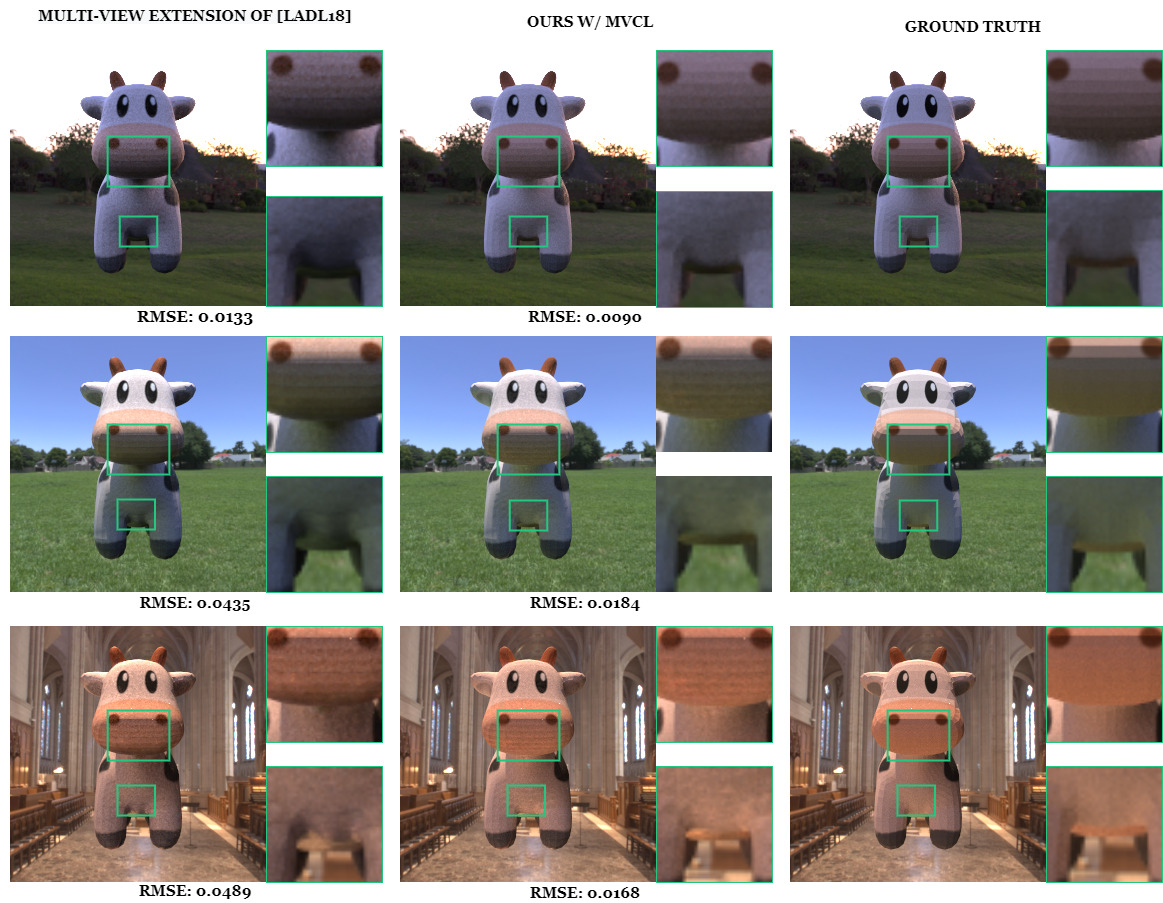}
\caption{Effects of entanglement. Our process achieves improved diffuse and specular disentanglement. The reflectance parameters are optimized under the light conditions of the top column. When the objects are re-rendered in novel lighting condition, the results of \cite{Li:2018:DMC} in left column shows a sharp increase in RMSE value, while it is less severe in our case shown in the middle column. }
\label{fig:effects_of_entaglement}
\end{figure}

We experimentally prove that our process achieves improved diffuse and specular reflectance disentanglement, compared with multi-view extension of \cite{Li:2018:DMC}. Using both techniques, we first optimize the parameters $\theta_{diff}, \theta_{spec}, \theta_{rough}, \theta_{env}$, and then re-render the object with the reflectance parameters $\theta_{diff}, \theta_{spec}, \theta_{rough}$ under a different environment map $\theta_{env}^{new}$, and calculate the RMSE of the reconstruction loss $\mathcal{L}_{recon}$ for a number of renders. Figure \ref{fig:effects_of_entaglement} shows the results for this experiment. Our technique produces the least amount of artifacts when the lighting conditions are changed. The inset images show close-ups of the same patches marked by the green rectangles. As shown, there is a sharp increase of the RMSE by 230\% using \cite{Li:2018:DMC} (first column) due to the severe entanglement in the reflectance, whereas only an 80-100\% increase of the RMSE using our technique, proving that in our case the entanglement of the diffuse and specular reflectance is improved.

\begin{figure*}[!ht]
\centering
\begin{subfigure}{\textwidth}
    \includegraphics[width=\textwidth]{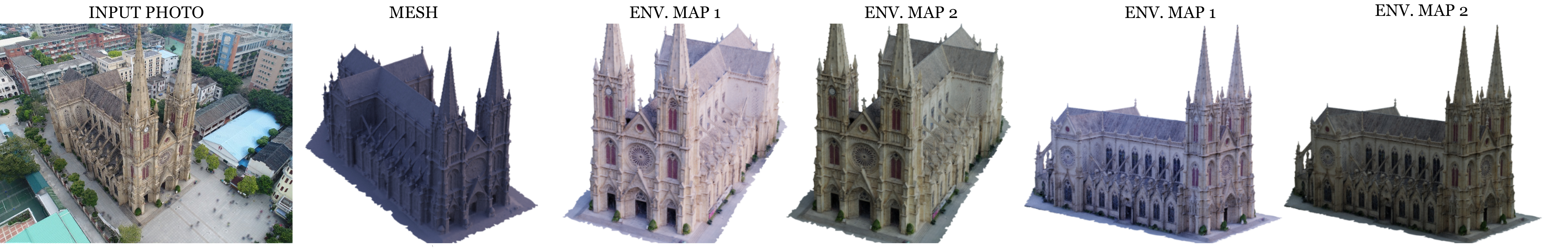}
    \caption{"Cathedral" scene (125 images, 454,854 triangles).}
    \label{fig:cathedral}
\end{subfigure}

\begin{subfigure}{\textwidth}
    \includegraphics[width=\textwidth]{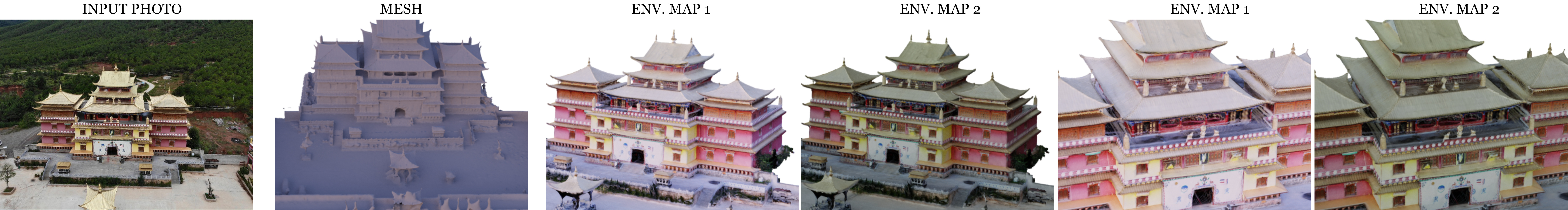}
    \caption{"Building" scene (132 images, 389,922 triangles).}
    \label{fig:}
\end{subfigure}

\begin{subfigure}{\textwidth}
    \includegraphics[width=\textwidth]{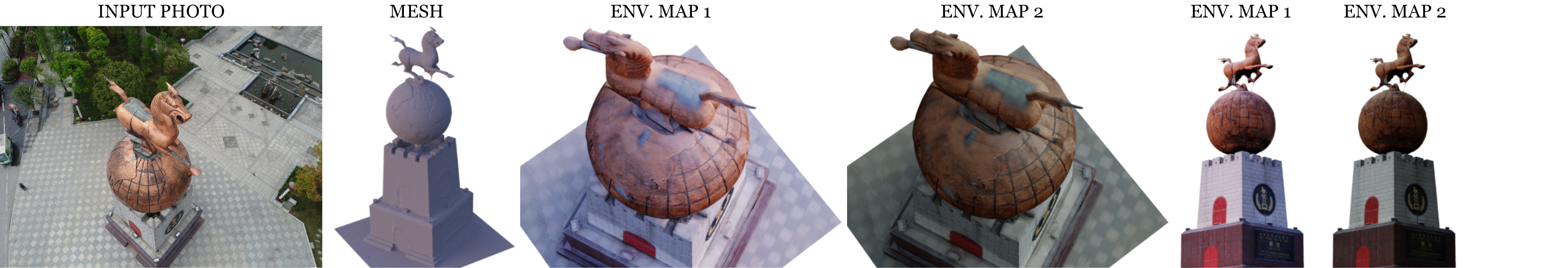}
    \caption{"Horse statue" scene (208 images, 152,298 triangles).}
    \label{fig:street_horse}
\end{subfigure}

\begin{subfigure}{\textwidth}
    \includegraphics[width=\textwidth]{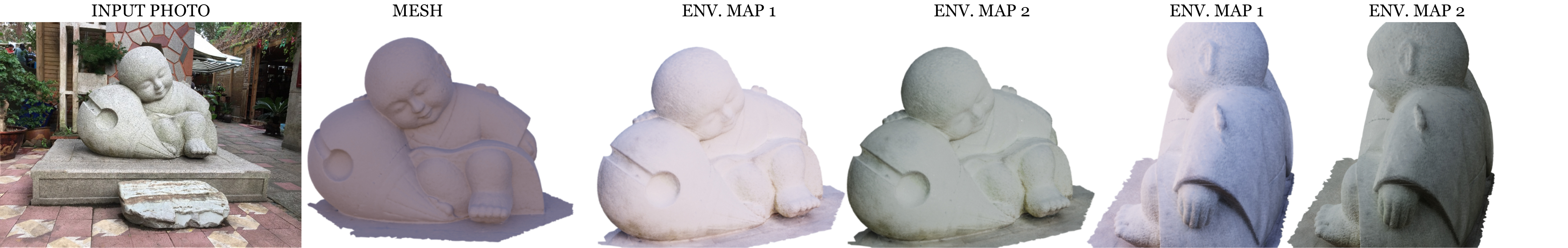}
    \caption{"Child statue" scene (148 images, 151,969 triangles).}
    \label{fig:}
\end{subfigure}

\begin{subfigure}{\textwidth}
    \includegraphics[width=\textwidth]{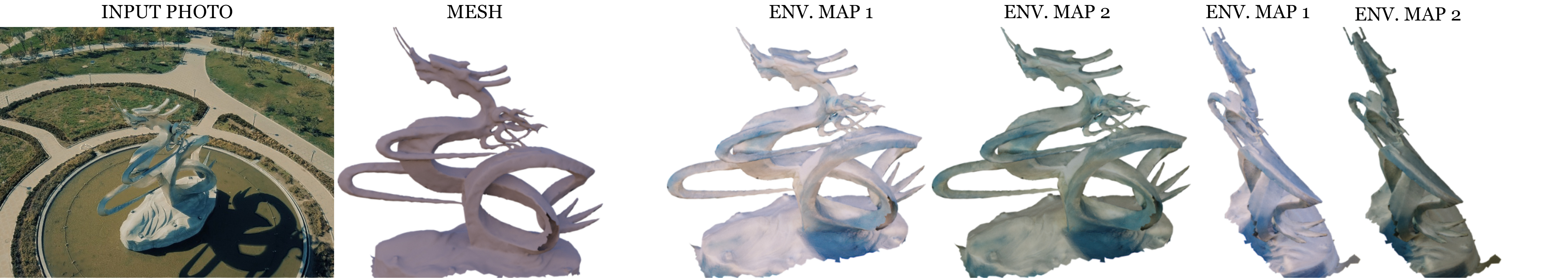}
    \caption{"Dragon statue" scene (338 images, 31,964 triangles).}
    \label{fig:dragon}
\end{subfigure}

\caption{Results of re-rendering real-world structures with the optimized reflectance. Our technique recovers detailed reflectance properties, providing realistic results under novel viewing and lighting conditions. Relighting is performed using HDR environment maps corresponding to different times of the day (ENV.MAP 1: morning and ENV.MAP 2: evening).}
\label{fig:outdoor_scenes}
\end{figure*}

\begin{figure*}[!ht]
    \centering
    \begin{subfigure}{0.49\textwidth}
        \includegraphics[width=\textwidth]{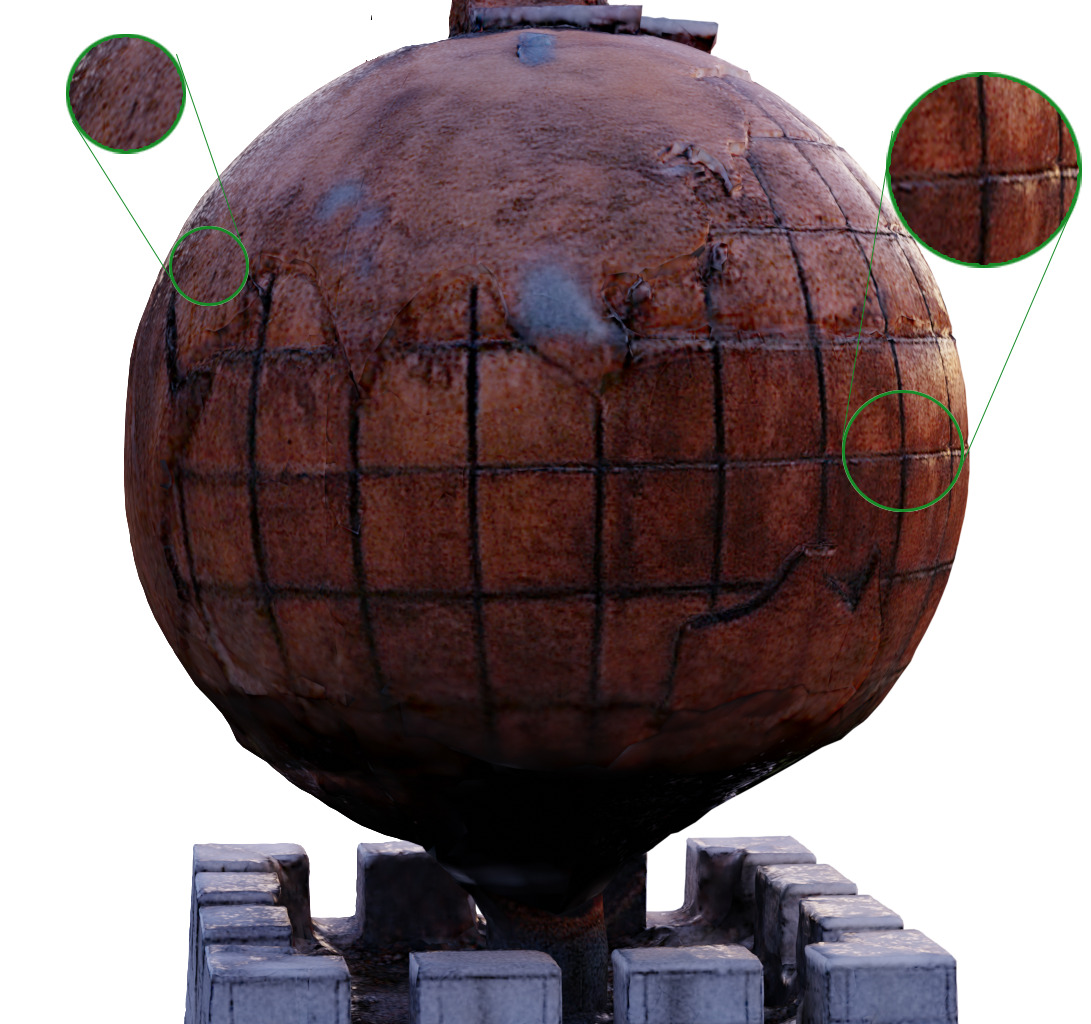}
        \caption{Well-lit environment map (morning).}
        \label{fig:globe_light}
    \end{subfigure}
    \begin{subfigure}{0.49\textwidth}
        \includegraphics[width=\textwidth]{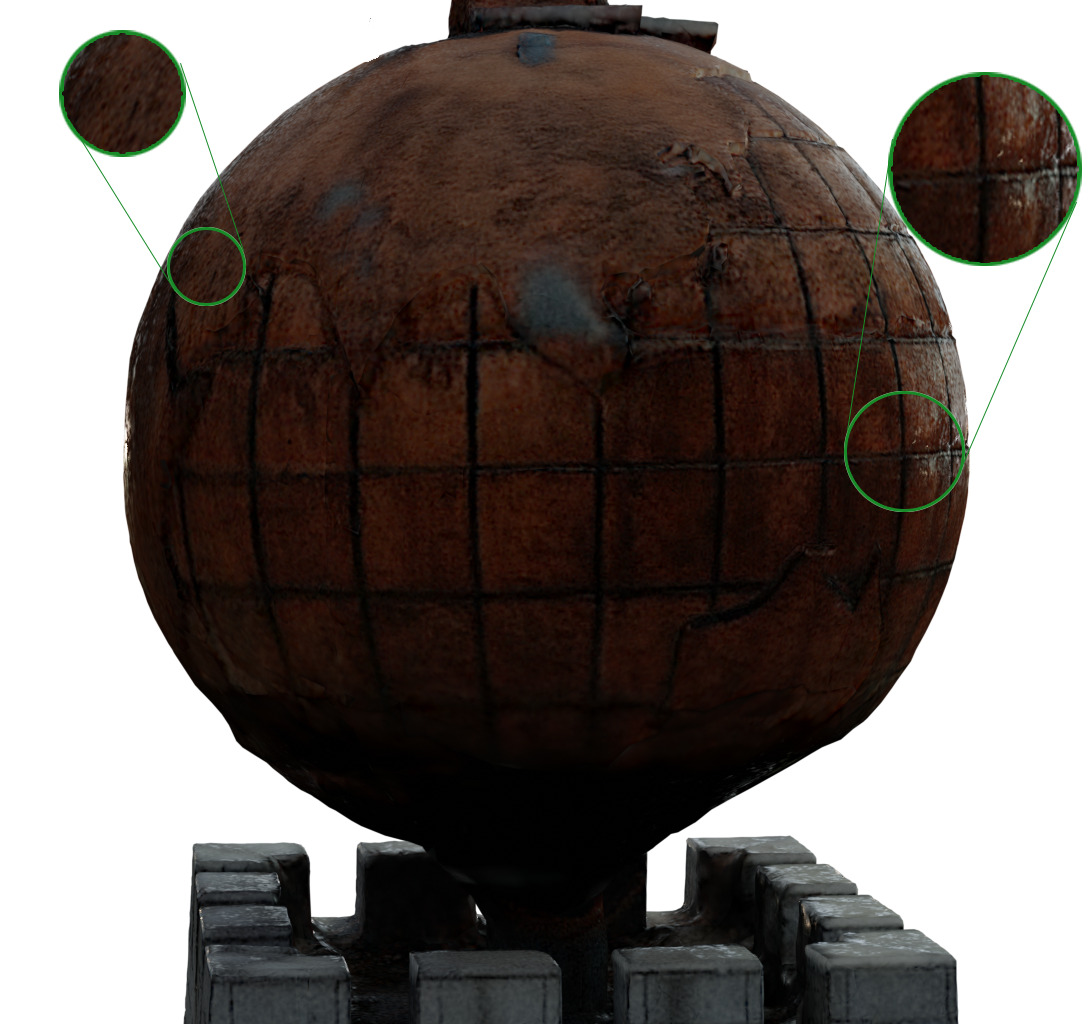}
        \caption{Low light (evening).}
        \label{fig:globe_dark}
    \end{subfigure}
    \caption{Specularity in outdoor scenes. Specularity in a scene is visible based on the intensity and direction of the incident. In a dimly lit environment, as in the left render, only minor specular highlights are observed, whereas, in a well-lit environment such as on the right, the colour and the specularity are easily noticeable. The sphere is a close-up of the horse statue shown in Figure \ref{fig:street_horse} and appears to be of metallic material i.e. copper.}
    \label{fig:horse_globe}
\end{figure*}

\begin{figure}[!ht]
    \centering
    \begin{subfigure}{0.5\textwidth}
        \includegraphics[width=\textwidth]{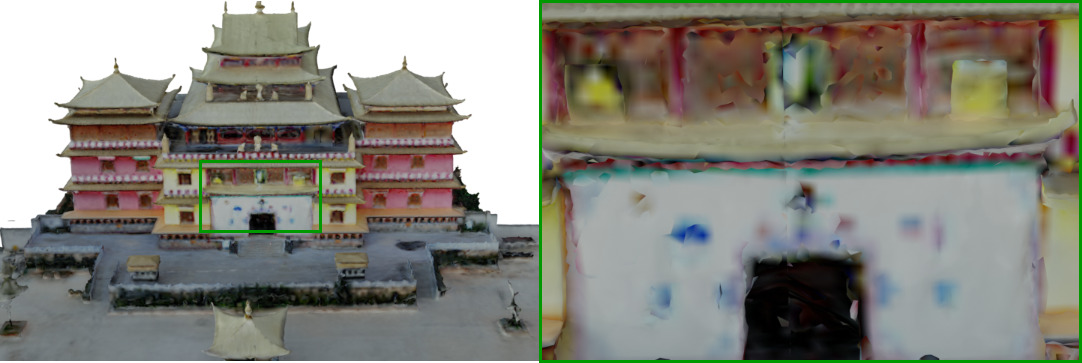}
        \caption{Rendered using 512x512 reflectance maps.}
        \label{fig:512}
    \end{subfigure}
    
    \begin{subfigure}{0.5\textwidth}
        \includegraphics[width=\textwidth]{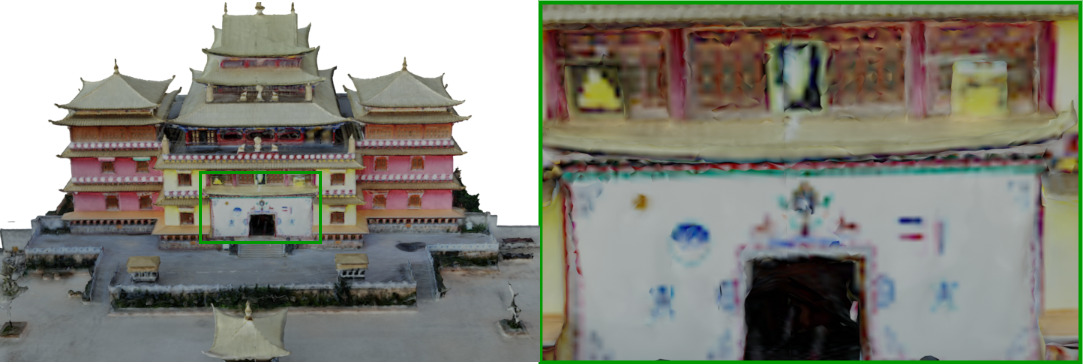}
        \caption{Rendered using 1024x1024 reflectance maps.}
        \label{fig:dragon}
    \end{subfigure}
    
    \begin{subfigure}{0.5\textwidth}
        \includegraphics[width=\textwidth]{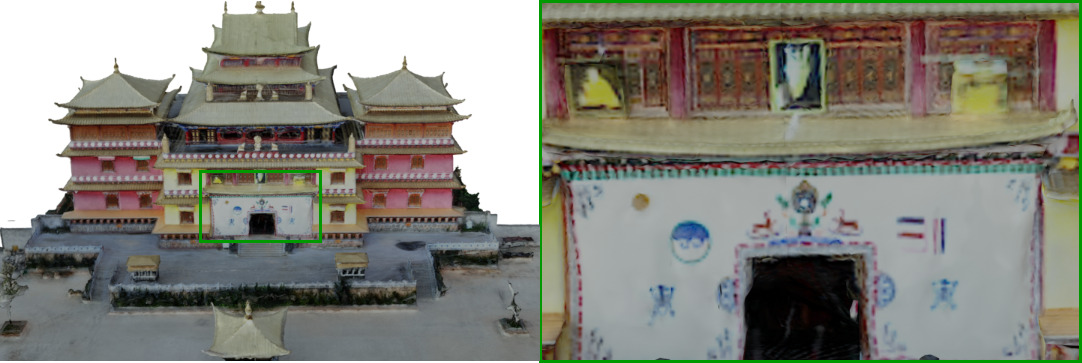}
        \caption{Rendered using 2048x2048 reflectance maps.}
        \label{fig:dragon}
    \end{subfigure}
    
    \begin{subfigure}{0.5\textwidth}
        \includegraphics[width=\textwidth]{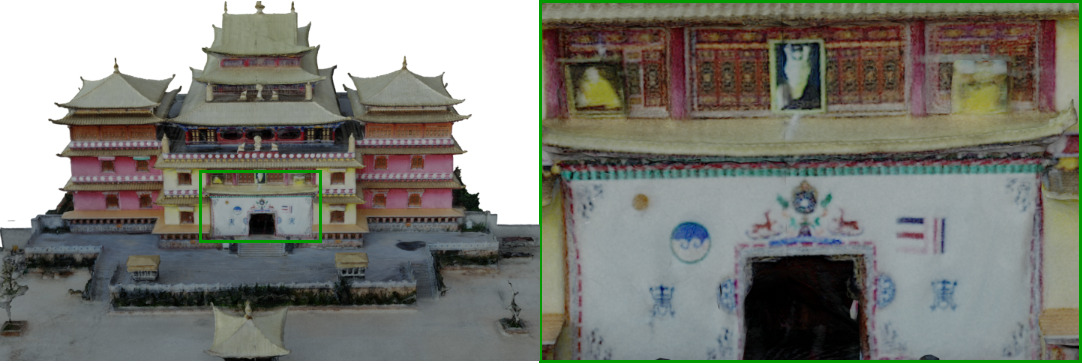}
        \caption{Rendered using 4096x4096 reflectance maps.}
        \label{fig:dragon}
    \end{subfigure}
    \caption{Effects of SVBRDF resolution. The use of higher resolution reflectance maps benefits the quality and realism of the renders. Using low resolution such as $512^2$ in large scale outdoor scenes causes blurry renders which can be avoided with higher resolutions including $2048^2$ and $4096^2$. A resolution of more than $2048^2$ provides negligible improvement on the loss at a significant computational cost. Renders on the left relit using ENV.MAP 2 (evening).}
    \label{fig:effects_of_resolution}
\end{figure}

\begin{figure}[!ht]
\centering
    \begin{subfigure}{0.23\textwidth}
        \includegraphics[width=\textwidth]{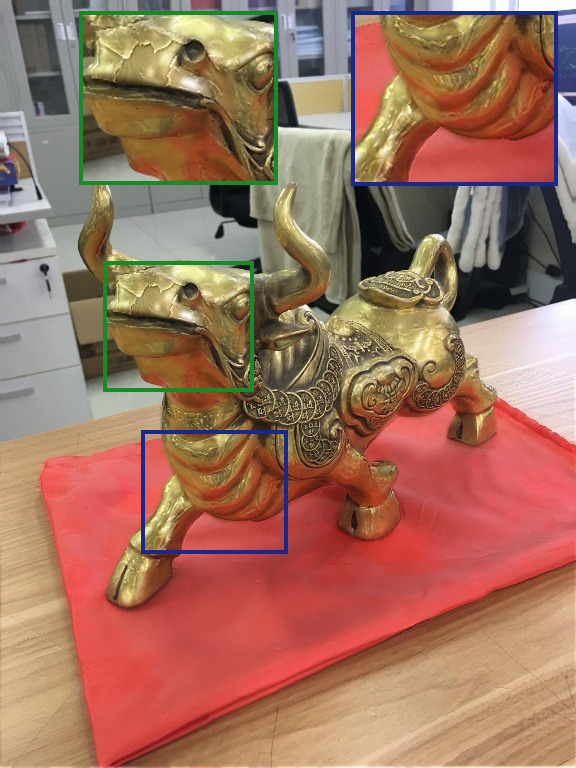}
        \caption{Sample input photo.}
        \label{}
    \end{subfigure}
     \begin{subfigure}{0.23\textwidth}
        \includegraphics[width=\textwidth]{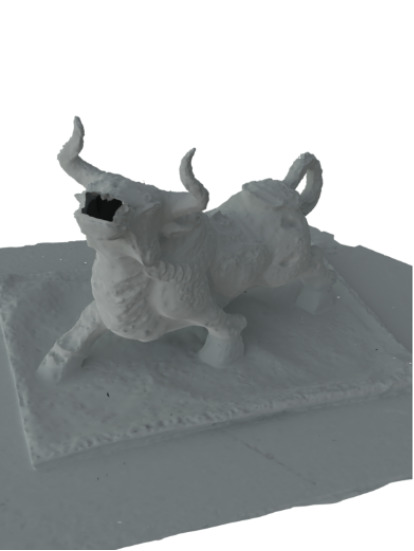}
        \caption{Scene's geometry.}
        \label{}
    \end{subfigure}
    
    \begin{subfigure}{0.23\textwidth}
        \includegraphics[width=\textwidth]{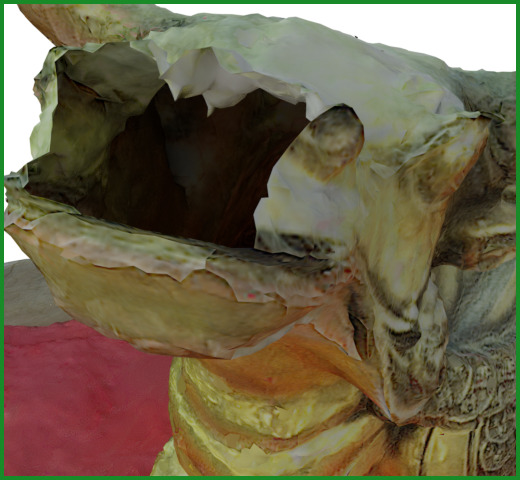}
        \caption{Rendered closeup of the nose.}
        \label{fig:render_closeup1}
    \end{subfigure}
     \begin{subfigure}{0.23\textwidth}
        \includegraphics[width=\textwidth]{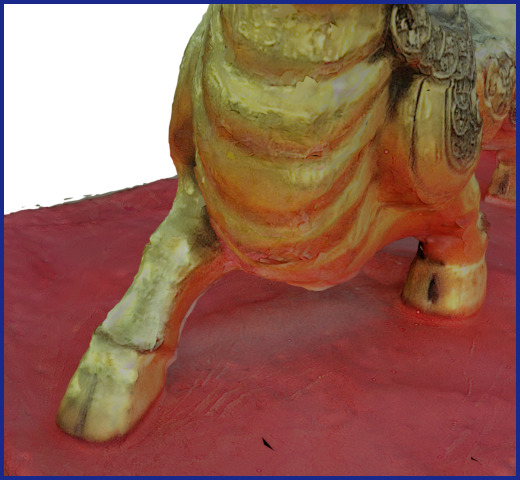}
        \caption{Rendered closeup of the neck.}
        \label{fig:render_closeup2}
    \end{subfigure}
 
\caption{Low-fidelity mesh. The recovery of reflectance properties can be negatively impacted by a low-fidelity mesh. Missing geometry around the nose leads to inaccuracies in the SVBRDFs in the local neighbourhood, as evident in (c). However, areas further away remain unaffected and achieve better results, as in (d).}
\label{fig:low_fidelity_mesh}
\end{figure}

\subsection{Complex outdoor scenes} 
We report the results of our experiments on the large-scale benchmark dataset BlendedMVS \cite{yao2020blendedmvs} which provides 20 to 1,000 multi-view images for 113 outdoor scenes. No lighting information or ground truth surface reflectance information is provided. The input data is generated from multi-view images of outdoor areas covering various scenes, including cities, architectures, sculptures and small objects under natural illumination conditions.  An online computer vision platform \cite{altizure} performs a complete 3D reconstruction and produces a 3D model representing the scene's geometry and the camera poses corresponding to each of the input images. We perform a degenerate dissolve on the scene's geometry to remove edges with no length and faces with no area. Calculating spatially varying reflectance requires the parameterization and mapping of the reflectance to the geometry. The reflectance properties are parameterized by diffuse, specular, and specular roughness maps that are subsequently mapped to the model using a quasi-conformal parameterization method based on a least-squares approximation of the Cauchy-Riemann equations \cite{levy2002conformal}. This process is robust to large textures atlases containing charts with complex borders and can handle disjoint objects with complex geometry when compared with other techniques such as angle-based flattening.

Figure \ref{fig:outdoor_scenes} shows the results of estimating reflectance from multi-view images of large-scale complex outdoor scenes. The scene's geometry is generated using multi-view stereo and given as input with the multi-view images, and associated camera poses. Reflectance is estimated using the proposed technique, and the scenes are then re-rendered under novel viewpoints and lighting conditions. We showcase a varying complexity of model geometries ranging from 31,964 triangles in \ref{fig:dragon} to 454,854 triangles in \ref{fig:cathedral}.

Figure \ref{fig:horse_globe} demonstrates a case of specularity evident from the reconstructed render using our technique from real-world images. The reflectance properties are properly estimated in this scenario to show accurate effects by showing highlights on the specular copper globe depending on the incident light direction. In Figure \ref{fig:globe_dark}, an evening environment map is used, where the intensity of the light is low, but the light comes off at an angle towards the object. This makes the object appear dark and shows small specularity in the edges as shown by the zoomed inset. For Figure \ref{fig:globe_light}, the model is rendered in a bright morning environment map. This shows the true colour of the model. The specularity is evident throughout and is not confined to small highlights as in the previous case.

\subsection{Effects of SVBRDF resolution}
We investigate the effects of the resolution for the SVBRDFs. Creating realistic renders requires high-fidelity reflectance estimates. This can be achieved with high-resolution resolution reflectance maps; however, at the cost of increased computational complexity and memory requirements. Figure \ref{fig:effects_of_resolution} shows an example of how the visual realism changes as a function of the reflectance map resolution used in the SVBRDF estimation. As expected, the lower resolutions, e.g. $512^2$ fail to capture essential details such as the ones appearing on the facade of the building. These hand-drawn markings shown in the closeups are distinguishable only when the resolution reaches $2048^2$. Visual improvements for $4096^2$ are trivial compared to the $2048^2$ and result in increased computational complexity and slower optimization convergence due to the increased memory requirements, which impose smaller batch sizes.

\section{Discussion and limitations} 
The proposed technique was extensively tested on numerous large-scale outdoor scenes with complex geometries and unknown natural illumination. Outdoor scenes mostly contain non-metallic surfaces, and our optimization technique and multi-view gradient consistency loss lead to significant improvement in terms of the reconstruction loss. Even in the presence of metallic surfaces, it outperforms the baseline, and we obtain high-quality results, as evident in Figure \ref{fig:horse_globe}. Since the reflection model cannot represent transparent or translucent materials, the SVBRDF of windows and glass-like materials cannot be accurately estimated.

The accuracy of the SVBRDF recovery relies on the quality of the input images and the generated scene's geometry. More fine-level details can be retrieved when higher resolution images are used. Similarly, the fidelity of the mesh can have detrimental effects on the accuracy of the optimization. When using meshes with low fidelity, the optimization performs gracefully, but inaccurately modelled areas have a negative impact on the estimates' accuracy. Figure \ref{fig:low_fidelity_mesh} shows an example of a metallic statue where the object's model is missing geometry from the nose of the bull due to insufficient coverage of the input images. In this case, the recovery of SVBRDF in the local neighbourhood around the nose area becomes less accurate than the rest of the body.




\section{Conclusion}
\label{sec:conclusion}
We presented an end-to-end process for estimating spatially varying surface reflectance of real-world scenes under natural illumination. We specifically focus on large-scale outdoor scenes with complex geometry and unknown illumination. We described the process which followed a two-step optimization and proposed a novel multi-view gradient consistency loss that minimizes the gradient variance and enforces the same per-point gradient direction. Experiments on synthetic scenes show that for non-metallic surfaces the RMSE is improved by up to 30-50\%, and for metallic surfaces up to 5-10\%. We further showed that the disentanglement of the diffuse and specular reflectance is improved and that re-lighting a scene with a new environment map results in a reduction of up to 50\% to the RMSE. Finally, we evaluated our technique on large-scale outdoor scenes with complex geometry and under unknown, natural illumination and presented our results. As shown, we recovered detailed SVBRDFs which provide realistic renders under novel viewing and lighting conditions. As part of our future work, we plan to extend our process to use BTDF to handle transparent or translucent materials such as windows and glass-like objects.

\ifCLASSOPTIONcompsoc
  \section*{Acknowledgments}
\else
  \section*{Acknowledgment}
\fi

This research is supported in part by the Natural Sciences and Engineering Research Council of Canada Grants DG-N01670 (Discovery Grant) and DND-N01885 (Collaborative Research and Development with the Department of National Defence Grant).

\ifCLASSOPTIONcaptionsoff
  \newpage
\fi



\bibliographystyle{abbrv-doi}
\bibliography{template}
%



%

\begin{IEEEbiography}[{\includegraphics[width = 1in,height = 1.2in,clip]{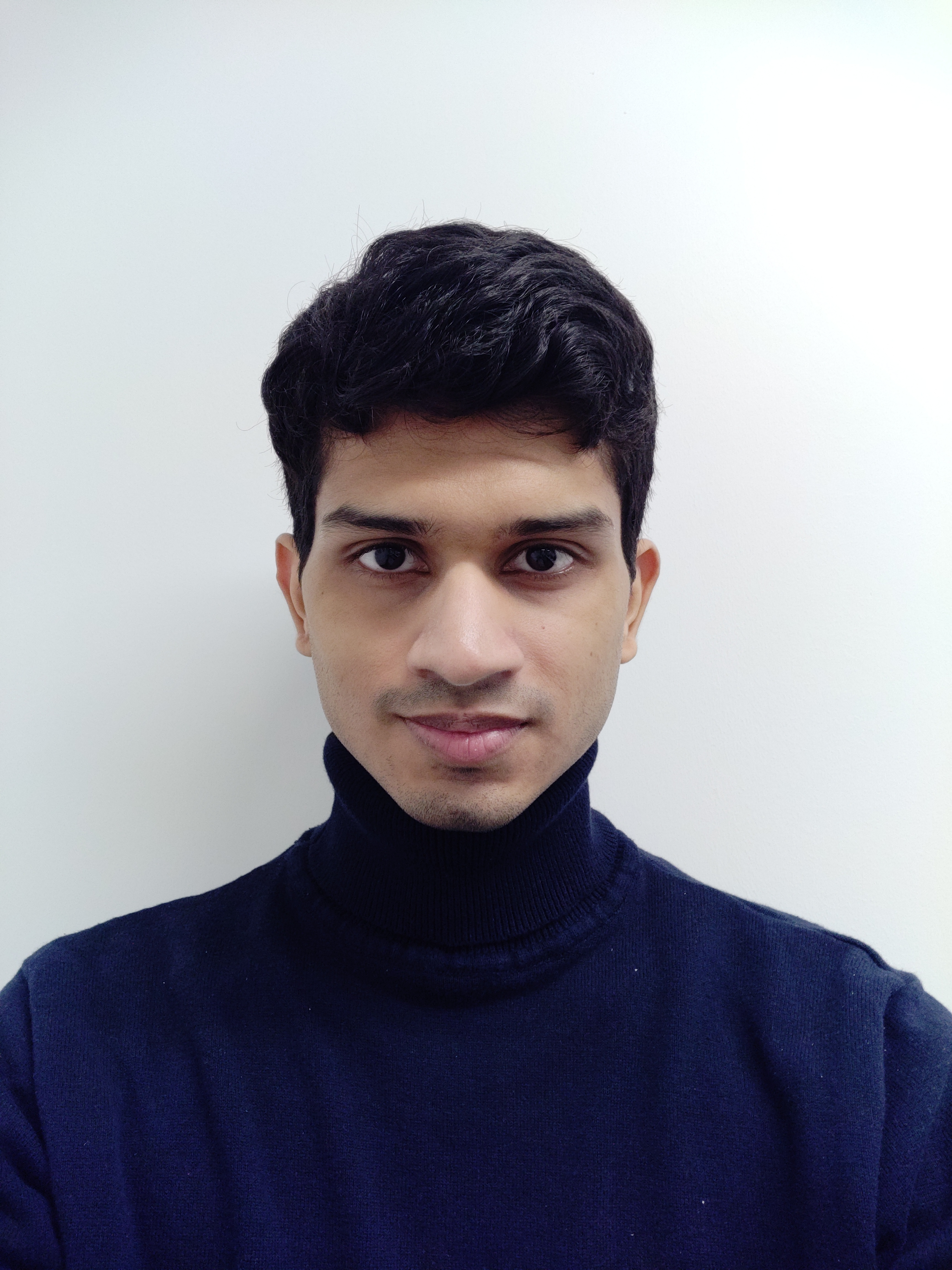}}]{Alen Joy}
Alen Joy is currently with Industrial Light \& Magic as an Associate Pipeline Technical Director.

Alen received his B.Tech in Computer Science and Engineering from TocH Institute Science and Technology, Kerala, India in 2018 and completed his M.Sc in Computer Science from Concordia University, Montréal, Canada in 2021.

He enjoys working on computer graphics and computer vision problems.
\end{IEEEbiography}

\begin{IEEEbiography}[{\includegraphics[width=1in,height=1in,clip,keepaspectratio]{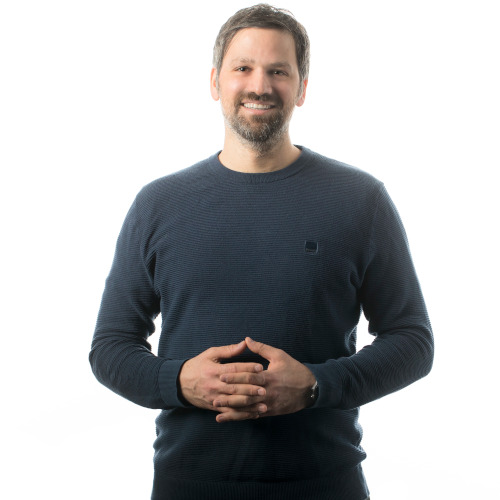}}]{Charalambos Poullis}
Charalambos (Charis) Poullis is an Associate Professor (Strategic Hire) at the Department of Computer Science and Software Engineering at the Gina Cody School of Engineering and Computer Science at Concordia University where he also serves as the Director of the Immersive and Creative Technologies (ICT) lab.

Charalambos received his B.Sc. in Computing and Information Systems with First Class Honors from the University of Manchester, UK, in 2001, an M.Sc. in Computer Science with specialisation in Multimedia and Creative Technologies (funded by a Fulbright-Amideast scholarship), and a Ph.D. in Computer Science (funded by USC) from the University of Southern California (USC), Los Angeles, USA, in 2003 and 2008, respectively.

His current research interests lie at the intersection of computer vision and computer graphics. More specifically, he conducts fundamental research in acquisition technologies \& 3D reconstruction, photo-realistic rendering, feature extraction \& classification; and applied research in virtual \& augmented reality.
\end{IEEEbiography}





\end{document}